\documentclass[pmlr,twocolumn,10pt]{jmlr} 
\usepackage{float}
\usepackage{caption}
\usepackage{booktabs}
\usepackage{siunitx}

\usepackage[switch]{lineno}

\usepackage{enum
item}       

\usepackage{multirow}

\usepackage{xcolor,colortbl}
\usepackage{color, colortbl} 
\usepackage{nicematrix}

\usepackage{xcolor} 

\definecolor{highlight}{rgb}{1.0,0.90,0.8}	
\definecolor{Gray}{gray}{0.96}

\definecolor{lightBlue}{rgb}{0.78, 0.85, 1.0}
\definecolor{lightOrange}{rgb}{0.88, 0.95, 1.0}
\definecolor{lightRed}{rgb}{1.0, 0.85, 0.85}
\usepackage{tcolorbox}
\usepackage{multirow}

\newtcbox{\bluebox}{on line, box align=base, colback=lightBlue,colframe=white,size=fbox,arc=3pt, before upper=\strut, top=-2pt, bottom=-4pt, left=-2pt, right=-2pt, boxrule=0pt}

\newtcbox{\orangebox}{on line, box align=base, colback=lightOrange,colframe=white,size=fbox,arc=3pt, before upper=\strut, top=-2pt, bottom=-4pt, left=-2pt, right=-2pt, boxrule=0pt}

\newtcbox{\redbox}{on line, box align=base, colback=lightRed,colframe=white,size=fbox,arc=3pt, before upper=\strut, top=-2pt, bottom=-4pt, left=-2pt, right=-2pt, boxrule=0pt}

\newtcbox{\whitebox}{on line, box align=base, colback=white,colframe=white,size=fbox,arc=3pt, before upper=\strut, top=-2pt, bottom=-4pt, left=-2pt, right=-2pt, boxrule=0pt}

\newcommand{\upshifted}{\raisebox{0.5\depth}{\tiny$\uparrow$}}

\newcommand{\uab}[1]{{\scriptsize\bluebox{\upshifted{#1}}}}

\newcommand{\equal}[1]{{\hypersetup{linkcolor=black}\thanks{#1}}}

\theorembodyfont{\upshape}
\theoremheaderfont{\scshape}
\theorempostheader{:}
\theoremsep{\newline}

\definecolor{highlight}{rgb}{1.0,0.90,0.8}	
\definecolor{Gray}{gray}{0.96}

\jmlrvolume{259}
\jmlryear{2024}
\jmlrsubmitted{LEAVE UNSET}
\jmlrpublished{LEAVE UNSET}
\jmlrworkshop{Machine Learning for Health (ML4H) 2024} 

\title[Multi-Modal SSL for Surgical Feedback Effectiveness Assessment]{Multi-Modal Self-Supervised Learning for Surgical Feedback Effectiveness Assessment}

\author{%
\Name{Arushi Gupta}\equal{These authors contributed equally}\Email{agupta5@caltech.edu}%
\AND
\Name{Rafal Kocielnik}\footnotemark[1] \Email{rafalko@caltech.edu}%
\AND
\Name{Jiayun Wang} \Email{peterw@caltech.edu}
\AND
\Name{Firdavs Nasriddinov} \Email{firdavs@caltech.edu}\\
\addr California Institute of Technology, USA
\AND
\Name{Cherine Yang
} \Email{cherine.yang@cshs.org}\\
\addr Cedars-Sinai Medical Center, USA
\AND
 \Name{Elyssa Wong} \Email{eywong@usc.edu}\\
\addr University of Southern California, USA
\AND
\Name{Anima Anandkumar} \Email{anima@caltech.edu}\\
\addr California Institute of Technology, USA
\AND
\Name{Andrew J. Hung} \Email{andrew.hung@cshs.org}\\
\addr Cedars-Sinai Medical Center, USA
}

\begin{document}

\maketitle

\begin{abstract}
During surgical training, real-time feedback from trainers to trainees is important for preventing errors and enhancing long-term skill acquisition. Accurately predicting the effectiveness of this feedback, specifically whether it leads to a change in trainee behavior, is crucial for developing methods for improving surgical training and education. However, relying on human annotations to assess feedback effectiveness is laborious and prone to biases, underscoring the need for an automated, scalable, and objective method. Creating such an automated system poses challenges, as it requires an understanding of both the verbal feedback delivered by the trainer and the visual context of the real-time surgical scene. To address this, we propose a method that integrates information from transcribed verbal feedback and corresponding surgical video to predict feedback effectiveness. Our findings show that both transcribed feedback and surgical video are individually predictive of trainee behavior changes, and their combination achieves an AUROC of $0.70\pm0.02$, improving prediction accuracy by up to 6.6\%. Additionally, we introduce self-supervised fine-tuning as a strategy for enhancing surgical video representation learning, which is scalable and further enhances prediction performance. Our results demonstrate the potential of multi-modal learning to advance the automated assessment of surgical feedback.
\end{abstract}


\paragraph*{Data and Code Availability}
Due to privacy restrictions, data is available upon request. The code is publically available \href{https://github.com/arushig100/Multi-Modal-SSL-for-Surgical-Feedback-Effectiveness-Assessment}{here}.

\paragraph*{Institutional Review Board (IRB)} The data used was collected under IRB of the University of Southern California (HS-17-00113).

\section{Introduction}
\label{sec:intro}

\paragraph{Importance:} 
During surgical training, real-time feedback from trainers to trainees with the intention of modifying trainee behavior is crucial for avoiding errors \citep{wong2023development} and enhancing long-term learning outcomes \citep{agha2015role}. 

\begin{figure*}[t!]
\centering
\includegraphics[width=\textwidth]{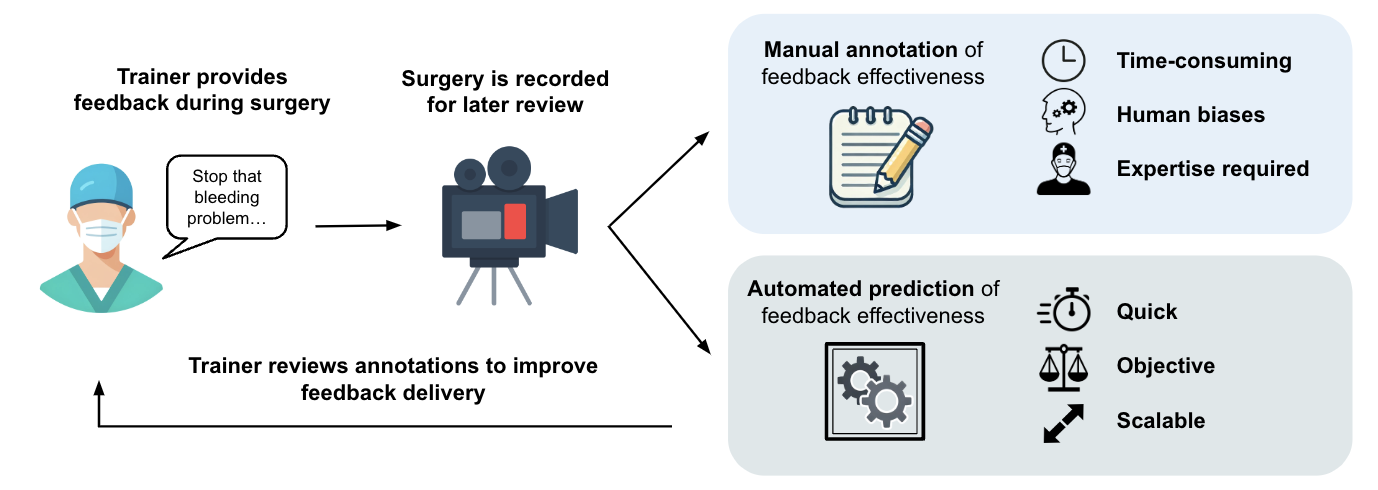}
\vspace{-12.0pt}
\caption{\textbf{Motivation for automating the prediction of feedback effectiveness in surgical training.} Manual annotation is time-consuming, prone to human cognitive biases (e.g., limited attention), and requires expertise, while automated prediction is quick, objective, and scalable. Trainers can use these assessments to improve their feedback delivery in the operating room, enhancing the quality of surgical training.}
\vspace{-4.0pt}
\phantomsection
\label{fig:motivation}
\end{figure*}

\begin{figure*}[t!]
\centering
\includegraphics[width=\textwidth]{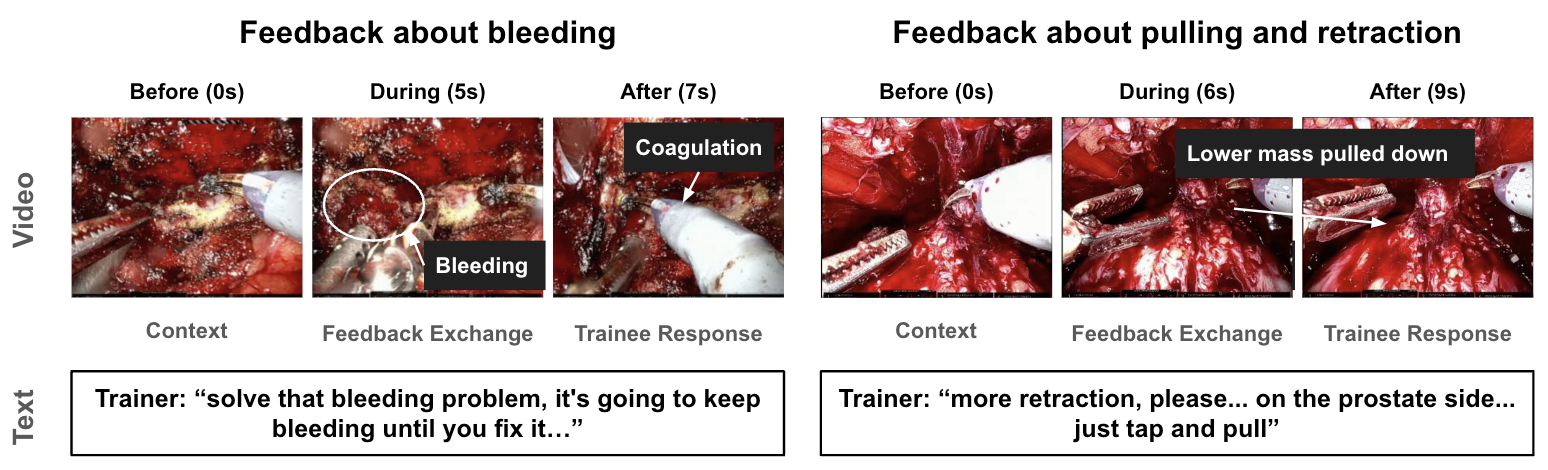}
\vspace{-12.0pt}
\caption{\textbf{Examples of feedback instances that are effective (result in trainee behavior change)}. The left example depicts feedback that describes addressing bleeding which causes the trainee to perform coagulation. The right example depicts feedback about pulling and retraction which causes the trainee to pull down the lower mass.}
\vspace{-4.0pt}
\label{fig:feedback_examples}
\phantomsection
\end{figure*}

The quality of such feedback has been shown to significantly influence a trainee's intraoperative performance \citep{bonrath2015comprehensive}, development of technical skills \citep{ma2022tailored}, and sense of autonomy \citep{haglund2021surgical}.
The quality of communication has also been directly linked to operative outcomes \citep{d2020evaluating}. Given its impact, understanding what makes feedback effective, specifically in terms of leading to trainee behavior change, is important. However, manually assessing feedback poses several challenges. First, it is inefficient, as human raters require at least twice the length of the surgical video to complete the assessment \citep{wong2023development}. Second, it is prone to human bias, as individual raters may be influenced by cognitive biases, such as fatigue or a limited attention span, and social biases, making consensus among multiple raters important for accuracy \citep{chinh2019ways}. Finally, it is resource-intensive as it requires the valuable time and expertise of skilled professionals.
As a result, there is a need for an objective, scalable, and automated approach for predicting feedback effectiveness (Figure \ref{fig:motivation}).

\paragraph{Challenges:} 
Analyzing feedback effectiveness is challenging due to its inherently unstructured and multimodal nature \citep{kocielnik2023deep, wong2023development}. Surgical feedback delivery involves a combination of verbal conversations and visual cues, both of which can vary significantly in form. Thus, evaluating feedback effectiveness requires a comprehensive understanding of the verbal feedback content and the visual surgical context \citep{wong2023development}. Feedback content provides information about the quality and nature of the trainer's guidance, while visual context provides insight into trainee behavior during feedback delivery, relevant complexities of the surgical scene (e.g., instrument usage, presence of bleeding, etc.), and post-feedback trainee behavior (Figure \ref{fig:feedback_examples}).

\begin{figure*}[ht]
\centering
\includegraphics[width=\textwidth]{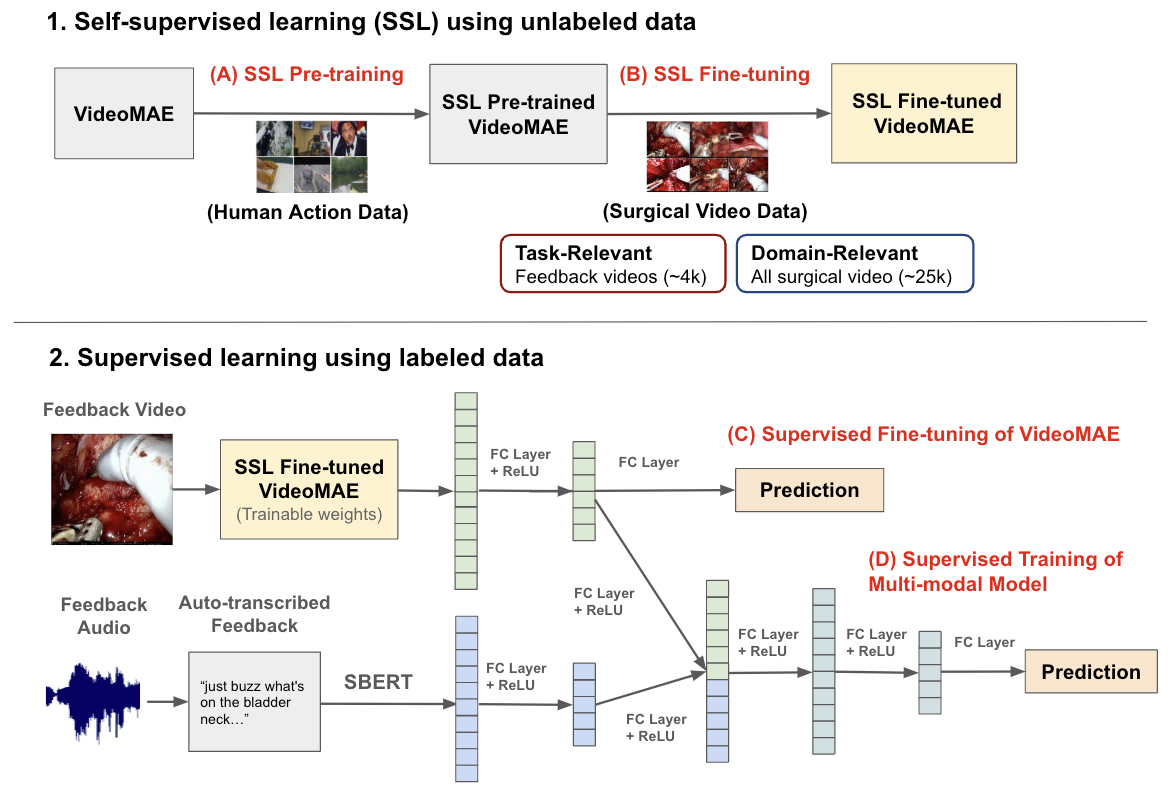}
\vspace{-12.0pt}
\caption{\textbf{The model pipeline} for predicting trainee behavior change from feedback audio and video, composed of four steps: \textbf{(A) Self-Supervised Pre-training} of VideoMAE using Kinetics-400 (human action data); \textbf{(B) Self-Supervised Fine-tuning} of VideoMAE using either \textbf{Task-Relevant} fine-tuning on feedback-specific videos or \textbf{Domain-Relevant} fine-tuning on all surgical video; \textbf{(C) Supervised Fine-tuning of VideoMAE} for predicting trainee behavior change; and \textbf{(D) Supervised Training of Multi-modal Model} which combines video features from VideoMAE and text features from SBERT for the prediction task.}
\label{fig:work_overview}
\vspace{-4.0pt}
\end{figure*}

\paragraph{Our Approach:}
In this work, we propose a multi-modal approach that integrates textual and visual information to predict feedback effectiveness. We transcribe audio recordings of feedback during surgery and use Sentence Transformers (SBERT) to extract embeddings from the transcriptions, capturing the semantic meaning of the trainer's guidance \citep{sbert2019}. On the visual side, we use video masked autoencoders (VideoMAE) to extract information from the surgical video corresponding to when feedback is delivered \citep{tong2022videomae}. To enhance model performance, we employ self-supervised learning (SSL) fine-tuning of VideoMAE on surgical data. We do SSL fine-tuning for the following reasons:
    \textbf{1) SSL is easily scalable:} SSL does not require labeled data, making it easy to continue SSL pre-training on additional domain-specific data as it is acquired \citep{caron2020unsupervised, kaplan2020scaling, sorscher2022beyond}.
    \textbf{2) Trainee behavior change is a weak label:} There is not always a clear link between surgical video corresponding to feedback and trainee behavior change, and even when there is, visual features indicative of trainee behavior vary across different feedback instances. SSL fine-tuning enables the model to learn more generalizable visual features from surgical video, which enhances the effectiveness of supervised training for the prediction task.    

We explore two SSL fine-tuning strategies: task-relevant SSL fine-tuning on task-relevant feedback video instances and domain-relevant SSL fine-tuning on all surgical videos. Both approaches result in improved predictive accuracy, while task-relevant SSL achieves comparable accuracy with 14.8\% of the data.

\paragraph{Findings:} 
We summarize our findings as follows:
\begin{itemize}[leftmargin=*, itemsep=0.0mm, topsep=5pt]
    \item We achieve an Area under the ROC Curve (AUC) of 0.70 $\pm$ 0.02 for the novel task of predicting trainee behavior change in surgical training.
    \item We show that self-supervised fine-tuning on surgical video data even with a limited amount of data significantly helps improve prediction performance.
    \item  We show that verbal feedback content and surgical video are predictive of trainee behavior, and the fusion of both improves prediction by up to 6.6\%. 
\end{itemize}

\paragraph{Contributions:}
\begin{itemize}[leftmargin=*, itemsep=0.0mm, topsep=5pt]
    \item To the best of our knowledge, we are the first to show the feasibility of an automated system to predict the effectiveness of surgical feedback in changing trainee behavior from multi-modal features of surgical video and audio.
    \item  We introduce self-supervised fine-tuning strategies to improve representation learning of surgical video.
    \item We provide novel insights into the importance of textual and visual features for evaluating feedback effectiveness in real-world surgeries.
\end{itemize}

\section{Related Work}
\label{sec:relatedw}
\paragraph{Feedback in Robot-Assisted Surgery.}
 Prior work on analyzing surgical feedback in robot-assisted surgery has primarily focused on categorization. \cite{wong2023development} most recently developed a clinically validated system to manually categorize feedback into three core components (Anatomic, Procedural, Technical) and auxiliary aspects (Praise, Criticism, Visual Aid). The study found some associations between these categories and trainee behavior change. However, its analysis was limited by the broadness of the categories, which do not capture the finer details necessary for a more accurate understanding of feedback effectiveness. Furthermore, the study relied heavily on manual effort to determine these categories, making the approach vulnerable to human limitations in recognizing complex patterns \citep{chan2021integrating}, inherent cognitive biases \citep{chinh2019ways}, and attention span constraints \citep{baralt2011coding}. 
 
 To the best of our knowledge, there exists no prior work on the automated prediction of feedback effectiveness from video and audio. 

\paragraph{Self-Supervised Video Understanding.}
The success of deep learning has relied heavily on the availability of large-scale annotated datasets, but obtaining these annotations is costly and labor-intensive, especially for surgical feedback, which additionally requires specialized expertise. Self-supervised learning (SSL) addresses this issue by enabling models to learn from unlabeled data \citep{schiappa2023self, tong2022videomae}. SSL trains models using objectives derived from the data itself, and the model is then later fine-tuned (using supervised learning) for specific downstream tasks \citep{schiappa2023self}. This approach is well-suited for feedback effectiveness prediction, where only a subset of surgery video data is tied to feedback instances.

A notable SSL technique for video is video masked autoencoders (VideoMAE), recognized for its data efficiency \citep{tong2022videomae}. VideoMAE masks parts of the video and trains the model to reconstruct missing segments, capturing essential contextual and temporal information \citep{tong2022videomae}.  VideoMAE pre-trained on human action data has been effective in surgical tasks like phase recognition in thoracic surgery \citep{mateen2024thoracic} and surgical feedback classification \citep{kocielnik2023deep}. We use VideoMAE to predict feedback effectiveness and explore further improvements through self-supervised fine-tuning.

\paragraph{Video/Multi-modal Models for Surgery.}
Most surgical computer vision research has focused on deep learning models for specific, highly visual tasks such as segmentation, tool identification, anatomic part recognition, and surgical scene reconstruction \citep{ramesh2023dissecting,twinanda2016endonet, allan20202018, jin2020multi, liu2024surgical, wang2022neural}. However, these models have mainly been tested on limited, procedure-specific datasets that do not fully capture real-world surgical workflows \citep{ramesh2023dissecting, schmidgall2024general}.

There has been limited progress in developing general or foundational surgical video models, as general models face challenges when applied to specific downstream tasks. For example, \cite{schmidgall2024general} introduced GSViT, a General Surgery Vision Transformer pre-trained on 680 hours of surgery videos using next-frame reconstruction. However, its focus on single-frame prediction limits its applicability for feedback prediction, where integrating information across multiple frames is key. \cite{yuan2023learning} proposed Surgical Vision Language Pre-training (SurgVLP) trained on 3k hours of surgical video lectures aligned with transcribed audio. While it outperformed non-surgical baselines, it underperformed compared to fully supervised methods on vision-specific surgical tasks, highlighting the difficulty of adapting general models to specialized tasks.

The only work exploring video use for surgical feedback tasks is \cite{kocielnik2023deep}, which developed a method for automated feedback classification using the clinically validated system proposed by \cite{wong2023development}. This approach utilized multi-modal inputs (text, audio, video) and fine-tuned a VideoMAE model pre-trained on human action data.

Although the tasks are similar, predicting feedback effectiveness is less visual compared to feedback classification, making it more challenging. Feedback categories such as ``Visual Aid'' and ``Anatomic'' involve distinct visual elements such as cues for directing the trainee's attention or the presence of anatomical landmarks \citep{wong2023development}. In contrast, understanding feedback effectiveness involves identifying unclear and variable visual cues \citep{kocielnik2023deep}.  

\section{Methods}
\label{sec:methods}
\subsection{Data and Annotations}
We use a dataset containing real-time feedback delivered during live robot-assisted surgeries summarized in Table \ref{tab:statistics}. It includes six types of surgeries: nephroureterectomy, inguinal hernia repair, partial nephrectomy, simple prostatectomy, nephrectomy, and radical prostatectomy. Each surgery was conducted using the da Vinci Xi robotic surgical system \citep{freschi2013technical}.
The dataset comprises audio recordings captured using wireless microphones and video footage from an endoscopic camera, both of which were simultaneously recorded during surgeries.  This feedback data was previously collected and categorized by \cite{wong2023development}.

A subset of the recorded OR conversations represents surgical feedback, defined by \cite{wong2023development} as trainer utterances intended to modify trainee behavior or thinking. Feedback instances were manually transcribed and timestamped by medical residents. The dataset encompasses 4204 distinct feedback instances across 33 surgeries.

For each feedback instance, the trainee's response was annotated into one of three categories: behavioral change, verbal acknowledgment, or request for clarification \citep{wong2023development}. Behavioral change was defined as an immediate adjustment made by the trainee in direct response to the feedback. Verbal acknowledgment involved the trainee confirming they had heard and understood the feedback. A request for clarification occurred when the trainee asked for the feedback to be restated due to a lack of understanding. Responses were annotated by the redundancy of 3 human raters and any rating discrepancies were discussed until a consensus was reached \citep{wong2023development}. 
\begin{table}[t!]
\centering
\caption{Statistics of video data and the annotations in our dataset per trainee response categories. Task-relevant SSL fine-tuning was performed on feedback video data, while domain-aware SSL fine-tuning was performed on all video data.}
\label{tab:statistics}
\begin{tabular}{@{}lrr@{}}
\toprule
\textbf{Category}              & \textbf{Len/Count} & \textbf{\% Total}       \\ \midrule
All Video                      & 78.9 hr       & 100\%                             \\
Feedback Video                 & 11.7 hr        & 14.8\%                            \\ \midrule
Feedback Instances   & 4204               & 100\%                             \\ \midrule
\textbf{Trainee Response}      &                    &                                    \\
Verbal Acknowledgment          & 1941               & 46.2\%                            \\
Behavioral Change              & 1865               & 44.4\%                            \\
Ask for Clarification          & 104                & 2.5\%                             \\ \midrule
\end{tabular}
\end{table}

Given that the primary goal of surgical feedback is to influence trainee behavior \citep{wong2023development}, we assess feedback effectiveness based on whether it results in trainee behavior change. 

\subsection{Model Pipeline}
The model pipeline for predicting trainee behavior consists of four key steps, as shown in Figure \ref{fig:work_overview}. First, VideoMAE is pre-trained using self-supervised learning (SSL) on an unlabeled human action dataset, Kinetics-400, as performed by \cite{tong2022videomae}. Next, we fine-tune VideoMAE with SSL on unlabeled surgical videos, using two strategies: task-relevant fine-tuning (using only feedback-related video) and domain-relevant fine-tuning (using all surgical video). After this, we perform supervised fine-tuning on labeled feedback videos to train the model specifically for predicting feedback effectiveness (trainee behavior change). Finally, we perform multi-modal supervised training which involves using both video features from VideoMAE and text features from SBERT to predict trainee behavior change. 

\subsection{Automated Speech Recognition}
We employed Automated Speech Recognition (ASR) to generate automated transcriptions based on the starting timestamps of feedback instances (for a 10s duration). For this, we used the pre-trained Whisper medium model \citep{radford2022whisper}, which was trained on 680,000 hours of labeled English speech data. Given the conversational nature of feedback exchanges, we also applied speaker diarization using Pyannote \citep{Bredin2020, Bredin2021}. Speaker diarization segments the audio to identify different speakers, adding context to the dialogue around feedback delivery. We extract features from the transcriptions using SBERT following \cite{kocielnik2023deep}.

\subsection{Self-Supervised Learning of VideoMAE}
 We used the VideoMAE base model, consisting of 86 million trainable parameters, pre-trained on Kinetics-400, a dataset comprising videos of 400 human action classes \citep{li2020ava}. We investigated task-relevant and domain-relevant SSL fine-training strategies. Task-relevant SSL fine-tuning involved pre-training the model on video clips associated with feedback instances, which comprised approximately 12 hours of data. Domain-relevant SSL fine-tuning involved pre-training on all available surgical videos, which comprised approximately 80 hours of data (Table \ref{tab:statistics}).

\subsection{Supervised Fine-tuning of VideoMAE}
Following SSL fine-tuning, the VideoMAE model was fine-tuned using supervised learning to predict feedback effectiveness (trainee behavior change). For each feedback video, features were extracted from VideoMAE and averaged across the temporal and spatial dimension, producing a 768-dimensional vector. This vector was then passed through three fully connected linear layers, reducing the dimensionality to 512, 256, and finally 2 (logits) in a funnel-shaped architecture. ReLU activations were applied between layers to introduce non-linearity.

\subsection{Supervised Training of Multi-modal Model}
Finally, features were extracted from VideoMAE (for video) and SBERT (for text) and integrated through supervised multi-modal training to predict trainee behavior change. Features extracted from VideoMAE were averaged across the temporal and spatial dimension resulting in a 768-dimensional vector. Two fully connected layers with ReLU activations were applied, reducing the dimensions to 512 and then 256. Features extracted from SBERT, initially a 384-dimensional vector, were also processed through two fully connected layers with ReLU activations, reducing their dimensions to 128 and then 64. The resulting vectors were concatenated into a 320-dimensional vector, which was passed through three fully connected layers, reducing its dimensions to 256, 128, and finally 2 (the logits) in a funnel architecture. During multi-modal training, SBERT and VideoMAE weights remained frozen.

\subsection{Data Processing}
Following \cite{kocielnik2023deep}, we trimmed a 10-second video when human-annotated feedback appears. This includes 5 seconds before (to capture context) and 5 seconds after (to capture delivery) the feedback onset. We preprocessed each video by downsampling the resolution to 320 x 250 and extracting 16 randomly uniformly sampled frames. 

\begin{table*}[ht]
\centering
\begin{tabular}{l | l l l}
\textbf{Method} & \textbf{AUROC} & \textbf{Precision} & \textbf{Recall} \\
\hline
Text & 0.66$_{\pm0.004}$ & 0.62$_{\pm0.02}$ & 0.62$_{\pm0.13}$ \\
Text + VideoMAE & 0.68$_{\pm0.01}\:$ \uab{3.59\%} & 0.63$_{\pm0.01}$ & 0.59$_{\pm0.10}$ \\
Text + VideoMAE (Task-relevant) & 0.70$_{\pm0.02}\:$ \uab{6.16\%} & 0.65$_{\pm0.03}$ & 0.56$_{\pm0.15}$ \\
Text + VideoMAE (Domain-relevant) & \textbf{0.70}$_{\pm0.02}$ \uab{6.55\%} & 0.63$_{\pm0.03}$ & 0.66$_{\pm0.09}$ \\
\hline
\end{tabular}
\vspace{-6.0pt}
\caption{Comparison of model performance for \textbf{predicting trainee behavior change} using auto-transcriptions of verbal feedback.  The percentage improvement shown is relative to the text method. Values represent the mean and standard deviation across three splits with different seeds. The multi-modal methods, combining video and text features, outperformed the text-only method, showing the value of incorporating visual information for the prediction task.}
\label{tab:auto_transcription_results}
\end{table*}

\begin{table*}[ht]
\centering
\begin{tabular}{l | l l l}
\textbf{Method} & \textbf{AUROC} & \textbf{Precision} & \textbf{Recall} \\
\hline
\rowcolor{Gray}SurgVLP & 0.52$_{\pm0.01}$ & 0.55$_{\pm0.06}$ & 0.42$_{\pm0.33}$\\
\rowcolor{Gray}GSViT & 0.50$_{\pm0.00}$ & 0.67$_{\pm0.29}$ & 0.49$_{\pm0.44}$\\
VideoMAE & 0.58$_{\pm0.00}$ & 0.55$_{\pm0.02}$ & 0.58$_{\pm0.37}$ \\
VideoMAE (Task-relevant) & \textbf{0.61}$_{\pm0.01}$ \uab{5.46\%} & 0.57$_{\pm0.04}$ & 0.61$_{\pm0.11}$ \\
VideoMAE (Domain-relevant) & 0.60$_{\pm0.01}\:$ \uab{3.73\%} & 0.56$_{\pm0.02}$ & 0.62$_{\pm0.22}$ \\
\hline

\end{tabular}
\vspace{-0.0pt}
\caption{Comparison of video-only model performance for \textbf{predicting trainee behavior change} across SurgVLP, GSViT, and multiple VideoMAE pre-training methods. Values represent the mean and standard deviation across three splits with different seeds. SurgVLP and GSViT, baseline models pre-trained on large surgical datasets and fine-tuned for this task, were not predictive of behavior change. VideoMAE features were somewhat predictive, with SSL fine-tuning of VideoMAE improving performance by up to 5.5\%.}
\vspace{-4pt}
\label{tab:video_results}
\end{table*}

\section{Experimental Results}
\label{sec:exp}
\subsection{Training and Implementation Details}
Task-relevant SSL fine-tuning was conducted for 500 epochs, and domain-relevant for 1000 epochs, both using a batch size of 24 and an initial learning rate of $1.5 \times 10^{-4}$. A linear decay scheduler reduced the learning rate to zero over training, with a weight decay of 0.05 applied. Following \cite{tong2022videomae}, which found that higher masking ratios result in better performance, an 85\% masking ratio was used for the reconstruction task.

Supervised fine-tuning of VideoMAE was conducted for 4 epochs with a batch size of 16, an initial learning rate of $1 \times 10^{-4}$, and a weight decay of 0.01. A ReduceLROnPlateau scheduler reduced the learning rate by 0.5 if validation accuracy didn't improve over two epochs. Multi-modal training was performed for another 4 epochs with a batch size of 16, a learning rate of $1 \times 10^{-3}$, and a weight decay of 0.01, with the learning rate managed similarly. 

All training was executed on an NVIDIA GeForce RTX 4090 GPU.

\subsection{Evaluation Protocols}
The dataset was split 80\% for training and 20\% for testing. To address the class imbalance in trainee behavior change, the minority class was upsampled in both sets.

In task-relevant SSL fine-tuning, all feedback instances excluding those in the test set were used. In domain-relevant SSL fine-tuning, all available surgical video data, split into non-overlapping 10-second clips, was used.

For all methods, hyperparameters, including epochs, learning rate, weight decay, and layer dimensions, were fine-tuned using grid search. This tuning was performed on a validation set comprising 12.5\% of the training data. Model performance was evaluated using Area Under the Receiver Operating Characteristic Curve (AUROC), with precision and recall metrics also reported. Methods were evaluated across three splits using distinct random seeds, with the mean and standard deviation reported.

\subsection{Results}
Table \ref{tab:auto_transcription_results} shows the performance of text-only and multi-modal methods using auto-transcribed text for predicting trainee behavior change. The text-only method achieved an AUROC of 0.66. Incorporating video features (without SSL fine-tuning) helped improve performance to an AUROC of 0.68, a 3.59\% improvement over the text baseline. With task-relevant and domain-relevant SSL pre-training, the AUROC improved to 0.70, reflecting gains of 6.16\% and 6.55\%, respectively. Both SSL fine-tuning approaches performed similarly, suggesting that using additional surgical video data did not yield significant benefits.

We further evaluated video-only methods to predict trainee behavior change as shown in Table \ref{tab:video_results}. Methods compared included SurgVLP \citep{yuan2023learning}, GSViT \citep{schmidgall2024general}, VideoMAE without SSL fine-tuning, and VideoMAE with task-relevant and domain-relevant SSL fine-tuning. SurgVLP and GSViT are existing surgery models pre-trained on large surgical datasets and serve as baselines (implementation details are provided in Appendix \ref{apd:baseline_video}). Overall, video features are less predictive than text, with the best method achieving an AUROC of 0.61. SurgVLP and GSViT were not predictive, with AUROCs of 0.52 and 0.50, respectively. Task-relevant and domain-relevant SSL fine-tuning of VideoMAE improved AUROC by 5.46\% and 3.73\% over VideoMAE without fine-tuning.

\subsection{Visual Inspection and Analysis}
Appendices  \ref{apd:pos_vis} and \ref{apd:neg_vis} provide examples of the multi-modal model's true positives, false positives, true negatives, and false negatives on the test data. Generally, in the cases where the model predicted a positive outcome, there was a noticeable change in the video during feedback delivery (at around 5 seconds) and after feedback delivery (at around 9 seconds). In contrast, in the cases where the model predicted a negative outcome, there was a minimal change.

False positive examples (Figure \ref{fig:tp_fp}) displayed substantial movement in the video that did not correspond to the verbal feedback. For instance, in the case of \textit{``Careful, you're almost cutting it,''} the instruments were moving, but the movement was unrelated to the feedback content. This may have caused the model to make a false positive prediction.

False negative examples (Figure \ref{fig:tn_fn}) displayed minimal movement, providing little visual evidence of a behavior change. For example, in the case of \textit{``You can clip that one,''} while the angle of the left instrument changed post-feedback, its position stayed the same. This lack of clear motion may have influenced the model to make a false negative prediction.

To gain further insight into model performance across various conditions, we analyze prediction accuracy by surgery type and by trainer ID (Appendix \ref{apd:accuracy_analysis}). The results show that while model performance is generally stable across different trainers and surgery types, accuracy may vary based on the number of instances and specific characteristics of each category.

Since it is impractical for surgeons to review full surgical videos, we additionally test using model logits as confidence scores to help prioritize the review of certain feedback segments (Appendix \ref{apd:confidence}). Predictions with higher confidence show greater accuracy, supporting this approach.

\section{Discussion}
\label{sec:discussion}

Our findings reveal the value of using multi-modal learning to assess feedback effectiveness in real-world surgeries. Incorporating features from video alongside auto-transcriptions improved performance by up to 6.6\%, resulting in an overall AUROC of 0.70.

\paragraph{Individual Predictiveness of Text and Video.}
We observe a significant difference in the individual predictiveness of the text and video modalities. The text-only model achieved an AUROC of 0.66, while the best video-only model achieved an AUROC of 0.61. This difference is likely because text is inherently more structured and less noisy compared to video data. The semantic meaning of the transcribed feedback is clearly relevant to the prediction task, whereas it is less clear which aspects of the video are relevant. Additionally, video data contains more noise (irrelevant information) due to its larger size and higher dimensionality, which makes it even more challenging to extract meaningful visual features for prediction with limited training data. 

\paragraph{SSL Fine-tuning.}
We demonstrate that self-supervised fine-tuning of video models effectively addresses the challenges of extracting meaningful information from video data. SSL task-relevant and domain-relevant fine-tuning improved video-only performance by up to 5.5\% and multi-modal performance by up to 6.6\% for the prediction task. SSL fine-tuning also offers scalability due to its reliance on unlabeled data \citep{caron2020unsupervised, kaplan2020scaling, sorscher2022beyond}. As more surgical video data is collected, it can be incorporated into the pre-training process, further boosting prediction performance and improving model adaptability over time. 

Interestingly, the task-relevant and domain-relevant fine-tuning methods performed comparably, suggesting that pre-training with additional video data beyond feedback instances was not significantly helpful. This may be because non-feedback video clips are (1) less likely to capture active surgical work that typically triggers feedback, and (2) less likely to show distinct visual changes seen in feedback clips, where trainee behavior often shifts visibly after feedback is delivered. So, the additional data may have been redundant or not as relevant \citep{sorscher2022beyond}. Although both methods achieve similar performance, they each have their advantages. Task-relevant SSL fine-tuning requires less data for the model to effectively learn task-relevant features \citep{sorscher2022beyond}.   Domain-relevant SSL fine-tuning does not require knowledge of the task, eliminating the need to determine which video instances are task-relevant. It is also more flexible for training \citep{oquab2023dinov2} and the resulting model may easily generalize to other tasks \citep{pantazis2021focus, tong2022videomae}.

\paragraph{Practical Value in Clinical Settings.}
Automating the prediction of feedback effectiveness can significantly improve surgical training. Compared to manual annotation, automated prediction is quick, standardized, and scalable \citep{wong2023development}.  Trainers can review predictions post-surgery and use them to improve their feedback delivery and training quality \citep{ma2024artificial}. Accurate feedback effectiveness assessment is also useful for the future development of systems for automated feedback delivery during live surgery, where assessing feedback options in real-time would be necessary \citep{kocielnik2023deep, ma2024artificial}. 

This work could be extended to improve training efficiency in additional ways.  For example, it could generate summaries for trainers highlighting which feedback features improved trainee performance and which were counterproductive, helping trainers refine feedback strategies. It could also provide drafts of evaluations from trainers to trainees.

\paragraph{Limitations and Future Directions.}
Our approach has inherent limitations due to data constraints. Since we do not know when feedback ends, we include 5 seconds of audio following each feedback starting timestamp to approximate capturing the full feedback delivery, but this may unintentionally include post-feedback speech. However, since explicit verbal commendations from trainers are relatively rare (14.8\% of instances), any potential data contamination is likely minimal and unlikely to significantly impact model performance \citep{wong2023development}.  Additionally, factors like trainee skill level introduce some inherent inter-individual variability (i.e., the same verbal feedback can result in different trainee responses). In cases where similar verbal feedback may occur, we account for the broader context by incorporating visual information into our predictions, making it unlikely for such variability to significantly affect model training. 

Although we show that the visual features extracted from VideoMAE are useful for predicting feedback effectiveness, incorporating visual features alongside auto-transcribed text features provides limited additional benefit to prediction performance. Trainee behavior changes are directly observable through video, indicating that we have yet to fully leverage the richness and relevant information present in video data, which will be the focus of future work. 

As a next step, while extracting general, abstracted visual features is valuable given the variability in visual cues, incorporating more structured visual information such as instrument kinematics and surgical gesture classification could further improve prediction performance. Another promising approach is to focus on identifying changes before and after feedback is delivered. For instance, contrastive learning could be applied to distinguish video segments pre- and post-feedback by creating positive pairs (related to the same feedback) and negative pairs (unrelated segments).

\section{Conclusion}
\label{sec:concl}
From a clinical perspective, both the content of feedback and the visual context play a crucial role in determining the effectiveness of surgical feedback during live surgeries. In this work, we demonstrate the value of a multi-modal approach, combining transcribed feedback and surgical video data, for the novel task of predicting feedback effectiveness. By accurately predicting feedback effectiveness in a scalable and objective way, we can gain a deeper understanding of what constitutes effective feedback and the visual contexts in which feedback is most or least effective. Trainers can use this information to provide more effective, personalized guidance to trainees, enhancing surgical training.  This understanding can also help pave the way for developing real-time systems that provide automated feedback during surgery. Ultimately, this work provides an important step towards advancing and automating surgical training in the operating room.

\acks{Research supported by NCI NIH under Award Number R01CA251579, R01CA273031 and the Northern California Associates SURF Fellowship. Content is solely the responsibility of authors and does not necessarily represent official views of NIH.}
\newpage
\bibliography{references}

\begin{thebibliography}{33}
\providecommand{\natexlab}[1]{#1}
\providecommand{\url}[1]{\texttt{#1}}
\expandafter\ifx\csname urlstyle\endcsname\relax
  \providecommand{\doi}[1]{doi: #1}\else
  \providecommand{\doi}{doi: \begingroup \urlstyle{rm}\Url}\fi

\bibitem[Agha et~al.(2015)Agha, Fowler, and Sevdalis]{agha2015role}
Riaz~A Agha, Alexander~J Fowler, and Nick Sevdalis.
\newblock The role of non-technical skills in surgery.
\newblock \emph{Annals of medicine and surgery}, 4\penalty0 (4):\penalty0 422--427, 2015.

\bibitem[Allan et~al.(2020)Allan, Kondo, Bodenstedt, Leger, Kadkhodamohammadi, Luengo, Fuentes, Flouty, Mohammed, Pedersen, et~al.]{allan20202018}
Max Allan, Satoshi Kondo, Sebastian Bodenstedt, Stefan Leger, Rahim Kadkhodamohammadi, Imanol Luengo, Felix Fuentes, Evangello Flouty, Ahmed Mohammed, Marius Pedersen, et~al.
\newblock 2018 robotic scene segmentation challenge.
\newblock \emph{arXiv preprint arXiv:2001.11190}, 2020.

\bibitem[Baralt(2011)]{baralt2011coding}
Melissa Baralt.
\newblock Coding qualitative data.
\newblock \emph{Research methods in second language acquisition: A practical guide}, pages 222--244, 2011.

\bibitem[Bonrath et~al.(2015)Bonrath, Dedy, Gordon, and Grantcharov]{bonrath2015comprehensive}
Esther~M Bonrath, Nicolas~J Dedy, Lauren~E Gordon, and Teodor~P Grantcharov.
\newblock Comprehensive surgical coaching enhances surgical skill in the operating room.
\newblock \emph{Annals of surgery}, 262\penalty0 (2):\penalty0 205--212, 2015.

\bibitem[{Bredin} and {Laurent}(2021)]{Bredin2021}
Herv{\'e} {Bredin} and Antoine {Laurent}.
\newblock {End-to-end speaker segmentation for overlap-aware resegmentation}.
\newblock In \emph{Proc. Interspeech 2021}, Brno, Czech Republic, August 2021.

\bibitem[{Bredin} et~al.(2020){Bredin}, {Yin}, {Coria}, {Gelly}, {Korshunov}, {Lavechin}, {Fustes}, {Titeux}, {Bouaziz}, and {Gill}]{Bredin2020}
Herv{\'e} {Bredin}, Ruiqing {Yin}, Juan~Manuel {Coria}, Gregory {Gelly}, Pavel {Korshunov}, Marvin {Lavechin}, Diego {Fustes}, Hadrien {Titeux}, Wassim {Bouaziz}, and Marie-Philippe {Gill}.
\newblock {pyannote.audio: neural building blocks for speaker diarization}.
\newblock In \emph{ICASSP 2020, IEEE International Conference on Acoustics, Speech, and Signal Processing}, Barcelona, Spain, May 2020.

\bibitem[Caron et~al.(2020)Caron, Misra, Mairal, Goyal, Bojanowski, and Joulin]{caron2020unsupervised}
Mathilde Caron, Ishan Misra, Julien Mairal, Priya Goyal, Piotr Bojanowski, and Armand Joulin.
\newblock Unsupervised learning of visual features by contrasting cluster assignments.
\newblock \emph{Advances in neural information processing systems}, 33:\penalty0 9912--9924, 2020.

\bibitem[Chan and Mazzocco(2021)]{chan2021integrating}
Jenny Yun-Chen Chan and Mich{\`e}le~MM Mazzocco.
\newblock Integrating qualitative and quantitative methods to develop a comprehensive coding manual: Measuring attention to mathematics in play contexts.
\newblock \emph{Methods in Psychology}, 4:\penalty0 100044, 2021.

\bibitem[Chinh et~al.(2019)Chinh, Zade, Ganji, and Aragon]{chinh2019ways}
Bonnie Chinh, Himanshu Zade, Abbas Ganji, and Cecilia Aragon.
\newblock Ways of qualitative coding: A case study of four strategies for resolving disagreements.
\newblock In \emph{Extended abstracts of the 2019 CHI conference on human factors in computing systems}, pages 1--6, 2019.

\bibitem[D’Angelo et~al.(2020)D’Angelo, Ruis, Collier, Shaffer, and Pugh]{d2020evaluating}
Anne-Lise~D D’Angelo, Andrew~R Ruis, Wesley Collier, David~Williamson Shaffer, and Carla~M Pugh.
\newblock Evaluating how residents talk and what it means for surgical performance in the simulation lab.
\newblock \emph{The American Journal of Surgery}, 220\penalty0 (1):\penalty0 37--43, 2020.

\bibitem[Freschi et~al.(2013)Freschi, Ferrari, Melfi, Ferrari, Mosca, and Cuschieri]{freschi2013technical}
Cinzia Freschi, Vincenzo Ferrari, Franca Melfi, Mauro Ferrari, Franco Mosca, and Alfred Cuschieri.
\newblock Technical review of the da vinci surgical telemanipulator.
\newblock \emph{The International Journal of Medical Robotics and Computer Assisted Surgery}, 9\penalty0 (4):\penalty0 396--406, 2013.

\bibitem[Haglund et~al.(2021)Haglund, Cutler, Suarez, Dharmapurikar, Lad, and McDaniel]{haglund2021surgical}
Michael~M Haglund, Andrew~B Cutler, Alexander Suarez, Rajeev Dharmapurikar, Shivanand~P Lad, and Katherine~E McDaniel.
\newblock The surgical autonomy program: a pilot study of social learning theory applied to competency-based neurosurgical education.
\newblock \emph{Neurosurgery}, 88\penalty0 (4):\penalty0 E345--E350, 2021.

\bibitem[Jin et~al.(2020)Jin, Li, Dou, Chen, Qin, Fu, and Heng]{jin2020multi}
Yueming Jin, Huaxia Li, Qi~Dou, Hao Chen, Jing Qin, Chi-Wing Fu, and Pheng-Ann Heng.
\newblock Multi-task recurrent convolutional network with correlation loss for surgical video analysis.
\newblock \emph{Medical image analysis}, 59:\penalty0 101572, 2020.

\bibitem[Kaplan et~al.(2020)Kaplan, McCandlish, Henighan, Brown, Chess, Child, Gray, Radford, Wu, and Amodei]{kaplan2020scaling}
Jared Kaplan, Sam McCandlish, Tom Henighan, Tom~B Brown, Benjamin Chess, Rewon Child, Scott Gray, Alec Radford, Jeffrey Wu, and Dario Amodei.
\newblock Scaling laws for neural language models.
\newblock \emph{arXiv preprint arXiv:2001.08361}, 2020.

\bibitem[Kocielnik et~al.(2023)Kocielnik, Wong, Chu, Lin, Huang, Wang, Anandkumar, and Hung]{kocielnik2023deep}
Rafal Kocielnik, Elyssa~Y Wong, Timothy~N Chu, Lydia Lin, De-An Huang, Jiayun Wang, Anima Anandkumar, and Andrew~J Hung.
\newblock Deep multimodal fusion for surgical feedback classification.
\newblock In \emph{Machine Learning for Health (ML4H)}, pages 256--267. PMLR, 2023.

\bibitem[Li et~al.(2020)Li, Thotakuri, Ross, Carreira, Vostrikov, and Zisserman]{li2020ava}
Ang Li, Meghana Thotakuri, David~A Ross, Jo{\~a}o Carreira, Alexander Vostrikov, and Andrew Zisserman.
\newblock The ava-kinetics localized human actions video dataset.
\newblock \emph{arXiv preprint arXiv:2005.00214}, 2020.

\bibitem[Liu et~al.(2024)Liu, Zhang, Wu, Hong, and Jin]{liu2024surgical}
Haofeng Liu, Erli Zhang, Junde Wu, Mingxuan Hong, and Yueming Jin.
\newblock Surgical sam 2: Real-time segment anything in surgical video by efficient frame pruning.
\newblock \emph{arXiv preprint arXiv:2408.07931}, 2024.

\bibitem[Ma et~al.(2022)Ma, Lee, Nguyen, Cowan, Haque, You, Roberts, Cen, Jarc, Gill, et~al.]{ma2022tailored}
Runzhuo Ma, Ryan~S Lee, Jessica~H Nguyen, Andrew Cowan, Taseen~F Haque, Jonathan You, Sidney~I Roberts, Steven Cen, Anthony Jarc, Inderbir~S Gill, et~al.
\newblock Tailored feedback based on clinically relevant performance metrics expedites the acquisition of robotic suturing skills—an unblinded pilot randomized controlled trial.
\newblock \emph{The Journal of Urology}, 208\penalty0 (2):\penalty0 414--424, 2022.

\bibitem[Ma et~al.(2024)Ma, Kiyasseh, Laca, Kocielnik, Wong, Chu, Cen, Yang, Dalieh, Haque, et~al.]{ma2024artificial}
Runzhuo Ma, Dani Kiyasseh, Jasper~A Laca, Rafal Kocielnik, Elyssa~Y Wong, Timothy~N Chu, Steven Cen, Cherine~H Yang, Istabraq~S Dalieh, Taseen~F Haque, et~al.
\newblock Artificial intelligence-based video feedback to improve novice performance on robotic suturing skills: a pilot study.
\newblock \emph{Journal of Endourology}, 2024.

\bibitem[Mateen et~al.(2024)Mateen, Malvia, Khader, Wang, Srinivasan, Yang, Schumacher, and Manjanna]{mateen2024thoracic}
Syed~Abdul Mateen, Niharika Malvia, Syed~Abdul Khader, Danny Wang, Deepti Srinivasan, Chi-Fu~Jeffrey Yang, Lana Schumacher, and Sandeep Manjanna.
\newblock Thoracic surgery video analysis for surgical phase recognition.
\newblock \emph{arXiv preprint arXiv:2406.09185}, 2024.

\bibitem[Oquab et~al.(2023)Oquab, Darcet, Moutakanni, Vo, Szafraniec, Khalidov, Fernandez, Haziza, Massa, El-Nouby, et~al.]{oquab2023dinov2}
Maxime Oquab, Timoth{\'e}e Darcet, Th{\'e}o Moutakanni, Huy Vo, Marc Szafraniec, Vasil Khalidov, Pierre Fernandez, Daniel Haziza, Francisco Massa, Alaaeldin El-Nouby, et~al.
\newblock Dinov2: Learning robust visual features without supervision.
\newblock \emph{arXiv preprint arXiv:2304.07193}, 2023.

\bibitem[Pantazis et~al.(2021)Pantazis, Brostow, Jones, and Mac~Aodha]{pantazis2021focus}
Omiros Pantazis, Gabriel~J Brostow, Kate~E Jones, and Oisin Mac~Aodha.
\newblock Focus on the positives: Self-supervised learning for biodiversity monitoring.
\newblock In \emph{Proceedings of the IEEE/CVF International conference on computer vision}, pages 10583--10592, 2021.

\bibitem[Radford et~al.(2022)Radford, Kim, Xu, Brockman, McLeavey, and Sutskever]{radford2022whisper}
Alec Radford, Jong~Wook Kim, Tao Xu, Greg Brockman, Christine McLeavey, and Ilya Sutskever.
\newblock Robust speech recognition via large-scale weak supervision, 2022.
\newblock URL \url{https://arxiv.org/abs/2212.04356}.

\bibitem[Ramesh et~al.(2023)Ramesh, Srivastav, Alapatt, Yu, Murali, Sestini, Nwoye, Hamoud, Sharma, Fleurentin, et~al.]{ramesh2023dissecting}
Sanat Ramesh, Vinkle Srivastav, Deepak Alapatt, Tong Yu, Aditya Murali, Luca Sestini, Chinedu~Innocent Nwoye, Idris Hamoud, Saurav Sharma, Antoine Fleurentin, et~al.
\newblock Dissecting self-supervised learning methods for surgical computer vision.
\newblock \emph{Medical Image Analysis}, 88:\penalty0 102844, 2023.

\bibitem[Reimers and Gurevych(2019)]{sbert2019}
Nils Reimers and Iryna Gurevych.
\newblock Sentence-bert: Sentence embeddings using siamese bert-networks.
\newblock \emph{CoRR}, abs/1908.10084, 2019.
\newblock URL \url{http://arxiv.org/abs/1908.10084}.

\bibitem[Schiappa et~al.(2023)Schiappa, Rawat, and Shah]{schiappa2023self}
Madeline~C Schiappa, Yogesh~S Rawat, and Mubarak Shah.
\newblock Self-supervised learning for videos: A survey.
\newblock \emph{ACM Computing Surveys}, 55\penalty0 (13s):\penalty0 1--37, 2023.

\bibitem[Schmidgall et~al.(2024)Schmidgall, Kim, Jopling, and Krieger]{schmidgall2024general}
Samuel Schmidgall, Ji~Woong Kim, Jeffery Jopling, and Axel Krieger.
\newblock General surgery vision transformer: A video pre-trained foundation model for general surgery.
\newblock \emph{arXiv preprint arXiv:2403.05949}, 2024.

\bibitem[Sorscher et~al.(2022)Sorscher, Geirhos, Shekhar, Ganguli, and Morcos]{sorscher2022beyond}
Ben Sorscher, Robert Geirhos, Shashank Shekhar, Surya Ganguli, and Ari Morcos.
\newblock Beyond neural scaling laws: beating power law scaling via data pruning.
\newblock \emph{Advances in Neural Information Processing Systems}, 35:\penalty0 19523--19536, 2022.

\bibitem[Tong et~al.(2022)Tong, Song, Wang, and Wang]{tong2022videomae}
Zhan Tong, Yibing Song, Jue Wang, and Limin Wang.
\newblock Video{MAE}: Masked autoencoders are data-efficient learners for self-supervised video pre-training.
\newblock In \emph{Advances in Neural Information Processing Systems}, 2022.

\bibitem[Twinanda et~al.(2016)Twinanda, Shehata, Mutter, Marescaux, De~Mathelin, and Padoy]{twinanda2016endonet}
Andru~P Twinanda, Sherif Shehata, Didier Mutter, Jacques Marescaux, Michel De~Mathelin, and Nicolas Padoy.
\newblock Endonet: a deep architecture for recognition tasks on laparoscopic videos.
\newblock \emph{IEEE transactions on medical imaging}, 36\penalty0 (1):\penalty0 86--97, 2016.

\bibitem[Wang et~al.(2022)Wang, Long, Fan, and Dou]{wang2022neural}
Yuehao Wang, Yonghao Long, Siu~Hin Fan, and Qi~Dou.
\newblock Neural rendering for stereo 3d reconstruction of deformable tissues in robotic surgery.
\newblock In \emph{International conference on medical image computing and computer-assisted intervention}, pages 431--441. Springer, 2022.

\bibitem[Wong et~al.(2023)Wong, Chu, Ma, Dalieh, Yang, Ramaswamy, Medina, Kocielnik, Ladi-Seyedian, Shtulman, et~al.]{wong2023development}
Elyssa~Y Wong, Timothy~N Chu, Runzhuo Ma, Istabraq~S Dalieh, Cherine~H Yang, Ashwin Ramaswamy, Luis~G Medina, Rafal Kocielnik, Seyedeh-Sanam Ladi-Seyedian, Andrew Shtulman, et~al.
\newblock Development of a classification system for live surgical feedback.
\newblock \emph{JAMA Network Open}, 6\penalty0 (6):\penalty0 e2320702--e2320702, 2023.

\bibitem[Yuan et~al.(2023)Yuan, Srivastav, Yu, Lavanchy, Mascagni, Navab, and Padoy]{yuan2023learning}
Kun Yuan, Vinkle Srivastav, Tong Yu, Joel Lavanchy, Pietro Mascagni, Nassir Navab, and Nicolas Padoy.
\newblock Learning multi-modal representations by watching hundreds of surgical video lectures.
\newblock \emph{arXiv preprint arXiv:2307.15220}, 2023.

\end{thebibliography}
\newpage
\appendix
\clearpage
\newpage

\section{True and False Positives Visualizations} \label{apd:pos_vis}
\begin{figure}[h]
\centering
\includegraphics[width=0.9\textwidth]{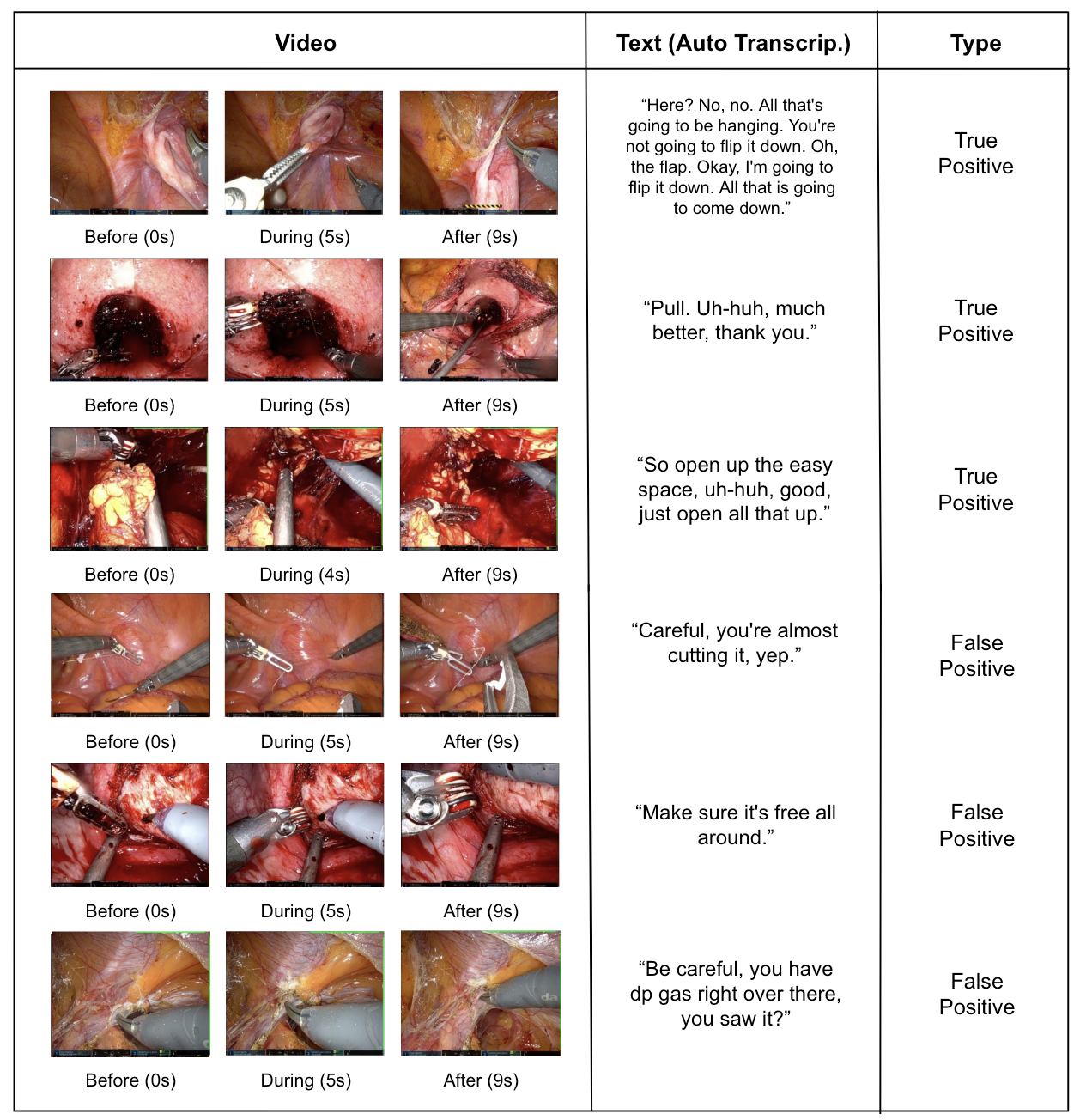}
\vspace{-12.0pt}
\caption{Examples of true positives and false positives using the multi-modal model for prediction of trainee behavior change.}
\vspace{-12.0pt}
\label{fig:tp_fp}
\end{figure}

\clearpage
\section{True and False Negatives Visualizations} \label{apd:neg_vis}
\begin{figure}[h]
\centering
\includegraphics[width=0.9\textwidth]{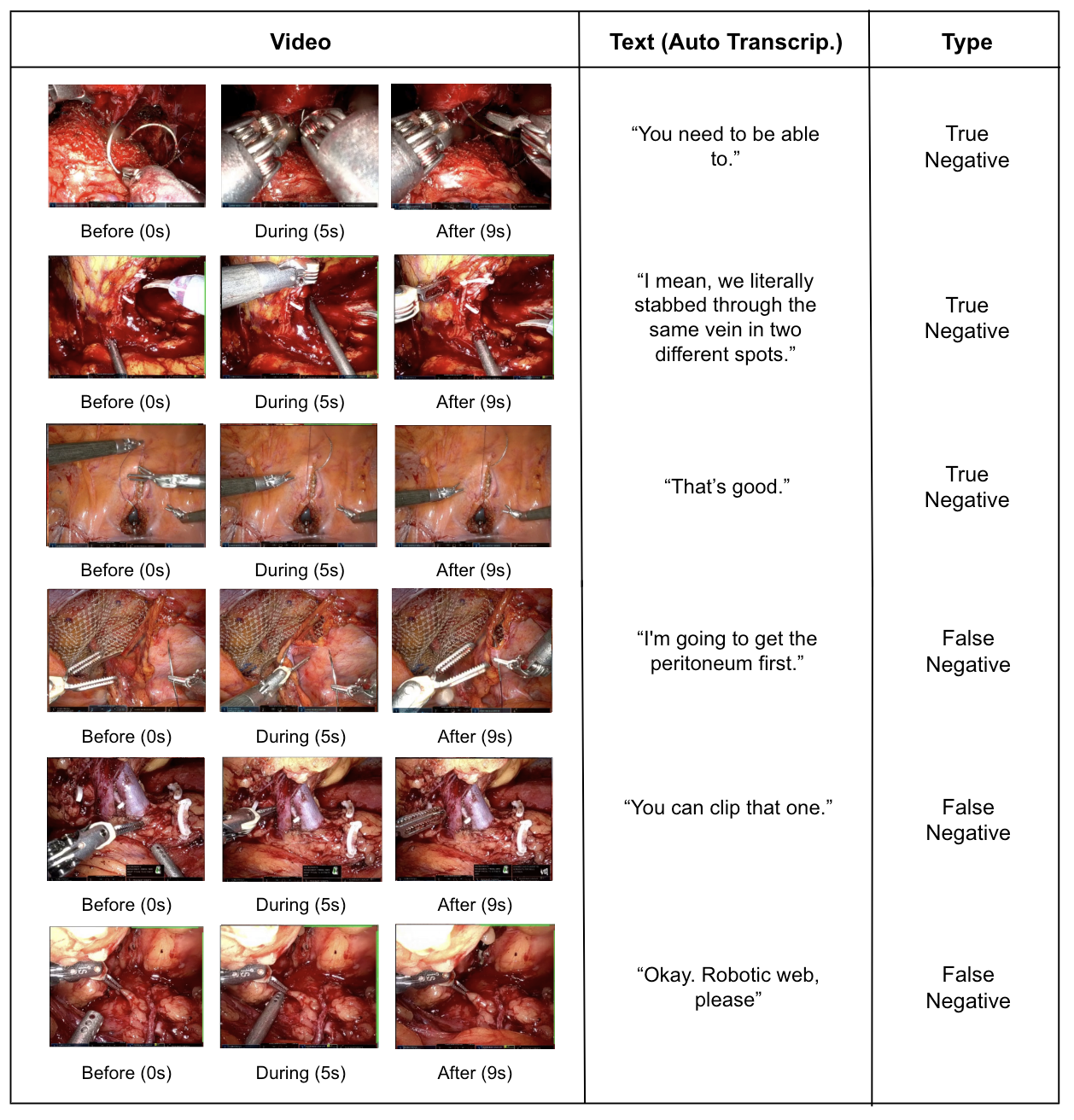}
\vspace{-12.0pt}
\caption{Examples of true negatives and false negatives using the multi-modal model for prediction of trainee behavior change.}
\vspace{-12.0pt}
\label{fig:tn_fn}
\end{figure}
\clearpage
\newpage

\section{Implementation Details of Baseline Video Models} \label{apd:baseline_video}
SurgVLP and GSViT were both fine-tuned for the task of feedback effectiveness prediction. The details of their implementations and fine-tuning are discussed below.
\\\\
\noindent \textbf{SurgVLP}: For each 10-second feedback video clip, 16 uniformly sampled frames were passed through the vision module of the SurgVLP model, producing a 16 × 768-dimensional output. This was averaged across the temporal dimension to generate a single 768-dimensional vector. A 2-layer multilayer perceptron (MLP) was then applied to obtain logits, and cross-entropy loss was minimized during training. The weights of the vision module were fine-tuned throughout the process. The model was trained for 5 epochs using the Adam optimizer with a learning rate of $1 \times 10^{-4}$.
\\\\
\noindent \textbf{GSViT}: For each 10-second feedback video clip, 16 preprocessed uniformly sampled frames were passed through GSViT to obtain individual embeddings, which were flattened and concatenated. A 3-layer MLP was then applied to obtain logits, and cross-entropy loss was minimized during training. GSViT weights were fine-tuned throughout the process. The model was trained for 10 epochs using the Adam optimizer with a learning rate of $1 \times 10^{-4}$.

\section{Analysis of Prediction Accuracy by Surgery Type and Trainer ID} \label{apd:accuracy_analysis}

To provide further insight into model performance, we present the F1 scores of predictions by surgery type and by trainer ID on the test set for one data split. Tables \ref{tab:surgery_type_f1} and \ref{tab:trainer_id_f1} show the breakdown of prediction accuracy across these categories.

\begin{table}[h]
\centering
\caption{Prediction F1 Scores by Surgery Type}
\label{tab:surgery_type_f1}
\begin{tabular}{ccc}
\hline
\textbf{Procedure}               & \textbf{$\#$ Instances} & \textbf{F1-bin} \\ \hline
Nephroureterectomy               & 6                            & 0.500          \\ \hline
Inguinal Hernia Repair           & 17                           & 0.632          \\ \hline
Partial Nephrectomy              & 41                           & 0.720          \\ \hline
Simple Prostatectomy             & 185                          & 0.648          \\ \hline
Nephrectomy                      & 173                          & 0.673          \\ \hline
Radical Prostatectomy            & 513                          & 0.597          \\ \hline
\end{tabular}
\end{table}

\begin{table}[h]
\centering
\caption{Prediction F1 Scores by Trainer ID}
\label{tab:trainer_id_f1}
\begin{tabular}{ccc}
\hline
\textbf{Trainer ID} & \textbf{$\#$ Instances} & \textbf{F1-bin} \\ \hline
A1                  & 296                          & 0.667          \\ \hline
A2                  & 465                          & 0.597          \\ \hline
A3                  & 114                          & 0.653          \\ \hline
A4                  & 60                           & 0.649          \\ \hline
\end{tabular}
\end{table}

The results show that while the model’s performance is generally stable across different trainers and surgery types, its accuracy may be influenced by the number of instances and specific characteristics of each category.

\section{Confidence Score Analysis} \label{apd:confidence}

To improve the practical utility of the model, we test using model logits as confidence scores to rank feedback instance predictions (Table \ref{tab:confidence_scores}). The results show that higher-confidence predictions are generally more accurate, suggesting that such a ranking can help surgeons prioritize which predictions to review. 

\begin{table}[H]
\centering
\caption{Model prediction accuracy at different confidence score thresholds}
\label{tab:confidence_scores}
\begin{tabular}{ccc}
\hline
\textbf{Confidence} & \textbf{$\%$ of Instances} & \textbf{Accuracy} \\ \hline
\textgreater 90\%         & 2.46\%                          & 87\%                          \\ \hline
\textgreater 85\%         & 6.53\%                          & 80\%                          \\ \hline
\textgreater 80\%         & 11.24\%                         & 76\%                          \\ \hline
\textgreater 75\%         & 22.59\%                         & 72\%                          \\ \hline
\textgreater 70\%         & 36.30\%                         & 70\%                          \\ \hline
\end{tabular}
\end{table}

\section{Segmented/Selective Manual Transcriptions} \label{apd:manual_transcriptions}

We investigated the performance of using manual transcriptions provided by the dataset for predicting feedback effectiveness. These manual transcriptions contain more information than simple verbatim transcriptions in two ways: 

\begin{enumerate}
    \item \textbf{Segmentation Based on Trainee Response:} Instead of transcribing a trainer’s spoken feedback as a continuous, uninterrupted utterance, human annotators split the feedback into smaller segments. Each segment corresponds to a part of the feedback that individually resulted in a trainee response such as verbal acknowledgment (which is inferred through audio) or behavior change (which is inferred visually). 
    
    \item \textbf{Selective Transcription:} In some cases, human annotators only transcribe parts of the trainer's spoken utterance that they believe are directly associated with an observed trainee response. 
\end{enumerate}

Examples of manual transcriptions illustrating these factors are shown in Table \ref{tab:transcription_comparison}. The extra judgment applied through segmentation and selective transcription makes manual transcriptions significantly more predictive of trainee behavior change, as shown in Table \ref{tab:human_transcription_results}. Since approximately half of the transcribed feedback instances resulted in trainee behavior change (as shown in Table \ref{tab:statistics}), the segmentation was partly informed by visual cues, meaning some video context is implicitly reflected in the manual transcriptions. This likely explains why the multi-modal approach offers only a modest performance improvement (up to 1.4\%) over the text-only approach when using manual transcriptions. In contrast, the multi-modal approach yields a larger performance improvement of up to 6.6\% when using auto-transcriptions.

\begin{table*}[ht]
\centering
\renewcommand{\arraystretch}{1.4} 
\begin{tabular}{|p{7cm}|p{7cm}|} 
\hline
\textbf{Utterance} & \textbf{Manual Transcription} \\
\hline
\multirow{6}{7cm}{
\raggedright
``That's not exactly it either. What I would suggest you do is I would suggest you grab this leading edge right there. Grab it. You're not doing right. No. No. This edge. The cut edge. Uh-huh. Grab it. Uh-huh. Good. Pull. Pull. Uh-huh. Open that up. But know where your vein is.''} & 
  ``that's not exactly it either.'' \\
\cline{2-2}
& ``what I would suggest you do is, is I would suggest you grab this leading edge right there, grab it.'' \\
\cline{2-2}
& ``this edge, the cut edge.'' \\
\cline{2-2}
& ``grab it.'' \\
\cline{2-2}
& ``good.'' \\
\cline{2-2}
& ``pull pull.'' \\
\cline{2-2}
& ``open that up.'' \\
\cline{2-2}
& ``wait, know where your vein is, know where your artery is.'' \\
\hline
\multirow{3}{7cm}{
\raggedright
``So take the... You're going to put only one clip to the artery. You're not going to cut the artery. Right. You want to be safe.''}
& ``only one clip to artery, wanna be safe.'' \\ & \\ & \\
\hline
\end{tabular}
\vspace{-6.0pt}
\caption{\textbf{Examples of manual transcriptions of feedback instances corresponding to continuous trainer utterances.} In the first example, the trainer's utterance is split into smaller segments, each resulting in a distinct trainee response (verbal acknowledgment or behavior change). In the second example, only the portions of the utterance directly relevant to the observed trainee response are transcribed (selective transcription).}
\label{tab:transcription_comparison}
\end{table*}

\begin{table*}[ht]
\centering
\begin{tabular}{l | l l l}
\textbf{Method} & \textbf{AUROC} & \textbf{Prec.} & \textbf{Recall} \\
\hline
Text & 0.79$_{0.04}$ & 0.72$_{0.03}$ & 0.72$_{0.11}$ \\
Text + Video & 0.80$_{0.03}$ \uab{0.53\%} & 0.74$_{0.03}$ & 0.68$_{0.06}$ \\
Text + Video (Task-Aware) & 0.80$_{0.03}$ \uab{1.36\%} & 0.73$_{0.04}$ & 0.71$_{0.02}$ \\
Text + Video (Domain-Aware) & 0.80$_{0.03}$ \uab{1.22\%} & 0.73$_{0.04}$ & 0.74$_{0.03}$ \\
\hline
\end{tabular}
\vspace{-6.0pt}
\caption{Comparison of model performance for \textbf{predicting trainee behavior change} using \textbf{manually transcribed text}. Each model is evaluated based on AUROC, precision, and recall. The percentage improvement shown is relative to the text method. Values represent the mean and standard deviation (SD) across three splits with different seeds.}
\label{tab:human_transcription_results}
\end{table*}

\end{document}